\documentclass[10pt,twocolumn,letterpaper]{article}

\usepackage[pagenumbers]{cvpr}      

\usepackage{graphicx}
\usepackage{amsmath}
\usepackage{amssymb}
\usepackage{booktabs}
\usepackage{subcaption}
\usepackage{xcolor}

\usepackage[pagebackref,breaklinks,colorlinks]{hyperref}

\usepackage[capitalize]{cleveref}
\crefname{section}{Sec.}{Secs.}
\Crefname{section}{Section}{Sections}
\Crefname{table}{Table}{Tables}
\crefname{table}{Tab.}{Tabs.}

\def\cvprPaperID{5584} 
\def\confName{CVPR}
\def\confYear{2022}

\begin{document}

\title{Visual Vibration Tomography: Estimating Interior \\ Material Properties from Monocular Video \vspace{-.2in}}

\author{Berthy T. Feng
\qquad Alexander C. Ogren
\qquad Chiara Daraio
\qquad Katherine L. Bouman
\\ California Institute of Technology
}

\twocolumn[{%
\renewcommand\twocolumn[1][]{#1}%
\maketitle
\begin{center}
  \centering
  \captionsetup{type=figure}
  \includegraphics[width=\textwidth]{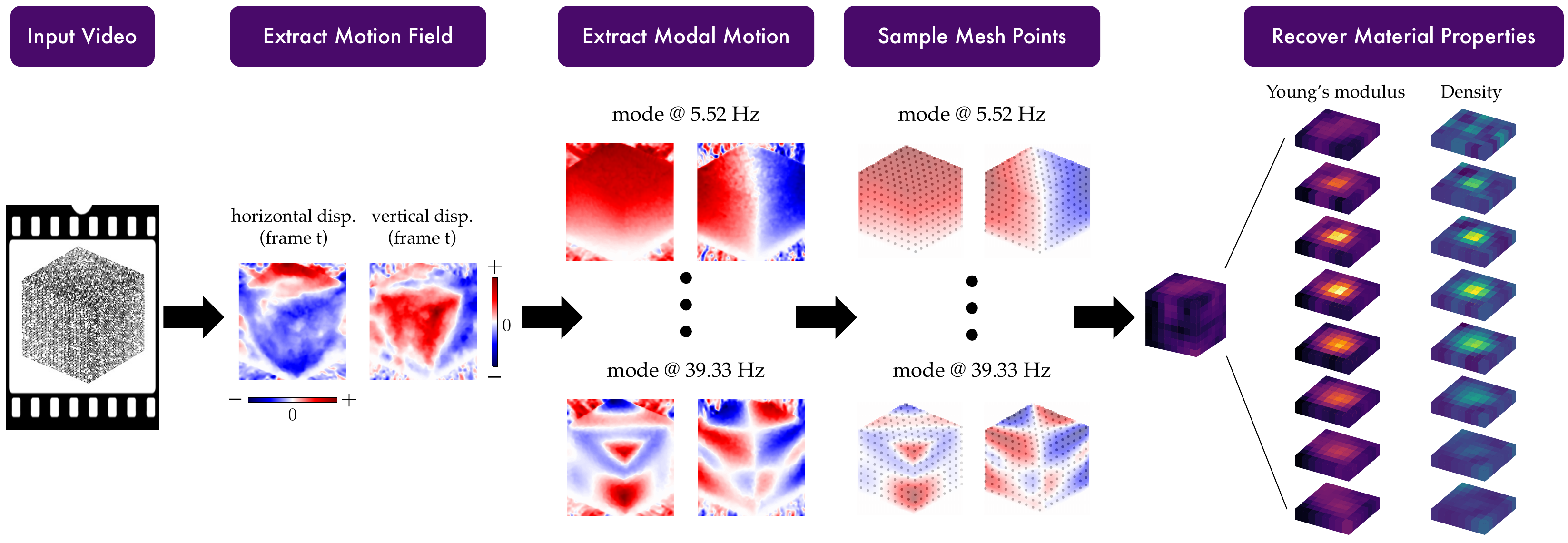}
  \captionof{figure}{Method overview. Starting with a video showing vibration of an object, we extract the motion fields across time and then decompose this motion into image-space modes.
  From the image-space modes sampled at visible mesh points, we are able to recover a voxelized volume of the Young's modulus and density throughout the interior of the object.}
  \label{fig:pipeline}
\end{center}%
}]

\begin{abstract}
\vspace{-.2in}
   An object's interior material properties, while invisible to the human eye, determine motion observed on its surface. We propose an approach that estimates heterogeneous material properties of an object from a monocular video of its surface vibrations. Specifically, we show how to estimate Young's modulus and density throughout a 3D object with known geometry. Knowledge of how these values change across the object is useful for simulating its motion and characterizing any defects. Traditional non-destructive testing approaches, which often require expensive instruments, generally estimate only homogenized material properties or simply identify the presence of defects. In contrast, our approach leverages monocular video to (1) identify image-space modes from an object's sub-pixel motion, and (2) directly infer spatially-varying Young's modulus and density values from the observed modes. We demonstrate our approach on both simulated and real videos.
\end{abstract}

\vspace{-7mm}
\section{Introduction}
\label{sec:intro}
\vspace{-.05in}
The subtle motions of objects around us are clues to their physical properties. Among such properties are stiffness and density, which dictate how an object will respond to environmental forces. As humans, we can vaguely characterize how stiff or heavy a material is, such as when we infer that a rubber basketball will bounce higher than a ceramic bowling ball by tapping on its surface. Most engineering applications, however, require a greater level of detail, such as when an aeronautical engineer must faithfully simulate how an airplane wing will react to wind turbulence. In computer vision and graphics, a full characterization of an object's material properties allows one to faithfully simulate its behavior. These scenarios require non-destructive testing to obtain physical properties of the object without altering it.

We propose \textit{visual vibration tomography}, a method to estimate material properties of an object directly from vibration signals extracted from monocular video. Much of non-destructive testing (NDT) has focused on measuring vibrations to identify the presence of defects in structures with a known geometry. However, NDT tools are not generally used to determine the precise spatial distribution of physical properties in objects with a heterogeneous interior structure.

We show that we can measure vibrations as sub-pixel motion in 2D video and then use this motion to constrain 3D material-property estimation. Videos have several advantages over existing NDT techniques: while contact sensors and laser vibrometers take point measurements, videos offer spatially dense  measurements of surface vibrations. While laser vibrometers are expensive and specialized, cameras are ubiquitous and general-purpose. While existing image-based techniques require stereo cameras for 3D motion tracking, our method shows that in many cases, a monocular view is all you need. 

Our motivating insight is that, under fixed geometry, an object's material properties determine its motion. The inverse direction is also true: motion determines material properties up to a scaling factor. If the motion is small, it can be decomposed into independent modes at natural frequencies, lending itself to a concise mathematical equation linking modes and material properties. This link lays the foundation for our physics-constrained optimization approach. The key challenge
of our task is to deal with incomplete and 2D (as opposed to full-field) modes. Despite these challenges, we show that we can
estimate material properties from image-space motion and recover
full-field modes.


In this paper, we first review related work and the theoretical relationship between modes and material properties. We then show how to extract image-space modes from video and recover material properties (Fig.~\ref{fig:pipeline} shows an overview of the method). We demonstrate our approach on simulated data of 3D geometries and discuss the effects of damping and model mismatch. Finally, we present proof-of-concept experiments on real data, showing that we are able to image the shape of unseen material inhomogeneities on drum heads and the presence of a defect in a real 3D Jello cube. These 
experiments demonstrate promise for the future of the approach in more challenging environments.
\footnote{Project website: {\scriptsize\url{http://imaging.cms.caltech.edu/vvt}}}

\section{Related Work}
\label{sec:related_work}

\subsection{Material Analysis from Images and Video} 
In computer vision, scene understanding is an important goal that includes, among many tasks, characterization of materials. Previous work has estimated material categories~\cite{liu2010exploring, bell2015material, Schwartz_2013_ICCV_Workshops} and surface properties~\cite{sharan2008image, ho2006direction} from images. 
In contrast to static images, videos have been used to estimate material properties, although these are often restricted to specific object categories, such as fabrics~\cite{bhat2003estimating, bickel2010design, Wang:2011:DDE, miguel2012data, bouman2013estimating} and trees~\cite{wang2017botanical}. Other work has inferred material properties from 3D point clouds~\cite{wang2015deformation, kim2017data} and known external forces~\cite{xu2015interactive}, but such measurements are harder to obtain than a 2D video. ``Visual vibrometry''~\cite{davis2015visual, visvib2017pami} uses a video's motion spectrum to estimate stiffness and damping of fabrics and rods. This is a promising step towards a general approach for estimating material properties, but it is restricted to homogenized properties. In a similar vein, others have used video data to identify structural modes~\cite{chen2014structural, YANG2020110183, s19051229}. Davis et al. demonstrated how to visualize image-space modes and use them for plausible simulation~\cite{davis2014visual, davis2015image}. 

\subsection{Traditional NDT}
Non-destructive testing (NDT) is an umbrella term for any technique that collects data of a material or structure without damaging it.
Usually, the goal is to identify defects or material inconsistencies that would change the expected behavior of the object. Laser vibrometry~\cite{durst1976principles} and digital image correlation (DIC)~\cite{chu1985applications} are popular tools for measuring surface displacements. 
Laser vibrometry has been used to examine the integrity of building structures~\cite{nassif2005comparison, roozen2015determining} and materials~\cite{emge2012remote, chen2014acoustic}.
DIC also has been used to identify defects in 2D structures~\cite{speranzini2014technique, tung2008development, feiteira2017monitoring, wu2011experimental, helm2008digital}.
Both laser vibrometry~\cite{macpherson2007multipoint, martarelli2001automated} and DIC~\cite{ha2015modal, trebuvna2014experimental} can be used for modal analysis, which involves identifying modal frequencies and shapes of a structure.
While usually regarded as a verification tool rather than a means
to directly infer material properties, recovered modal information has been used to solve for homogenized material properties~\cite{davis2015visual,foti2012ambient}. However, to 
our knowledge, modal analysis has not been used to solve the more challenging inverse problem of quantifying the heterogeneous properties addressed in this paper. 
\vspace{-.1in}
\section{Background}
\label{sec:background}
\subsection{Modal Analysis}
\label{subsec:modal_analysis}
\vspace{-.1in}

Every object has resonant, or natural, frequencies. At each {\it resonant
frequency}, the object vibrates periodically in a particular shape,
called a {\it mode}.
The vibration of a linear elastic object can be decomposed into independent modes. 

In the finite element method (FEM), we model an object as a mesh, composed of elements that each take on material-property values. The mechanical properties that determine an object's vibration are Young's modulus ($E$), Poisson's ratio ($\nu$), and density ($\rho$). $E$ and $\nu$ define the stiffness of connections between vertices, while $\rho$ defines the mass distribution. In this discretized model, the $n\times n$ stiffness matrix $K$ describes the stiffness between each pair of $n$ total DOFs, and the $n\times n$ mass matrix $M$ describes the mass concentrated between each pair of DOFs.
A mode $u$ and frequency $\omega$ are an eigenvector-eigenvalue solution of the generalized eigenvalue problem:
\begin{equation}
\label{eq:geip}
\setlength{\abovedisplayskip}{1pt}
\setlength{\belowdisplayskip}{1pt}
    K u  = \omega^2 M u.
\end{equation}

\begin{figure}[tb]
    \centering
    \includegraphics[width=0.3\textwidth]{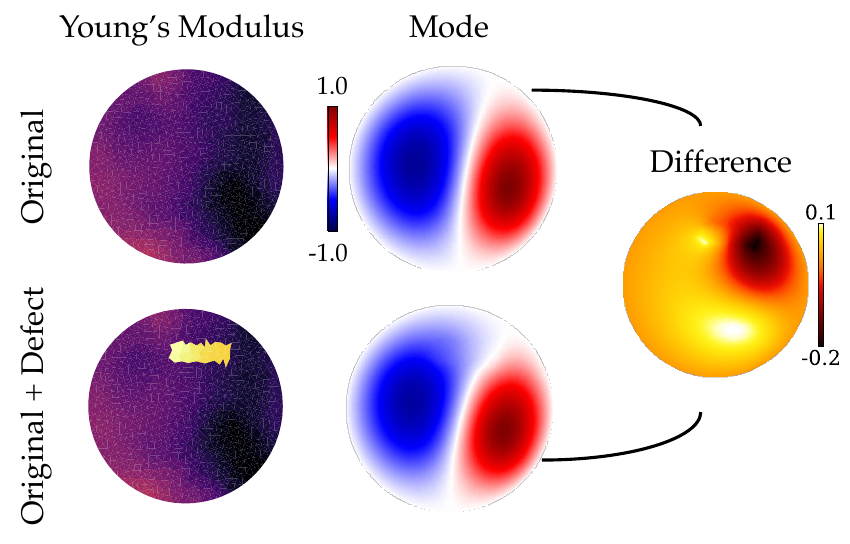}
    \caption{Small changes in material properties affect motion. Here a small region of a circular membrane becomes stiffer from ``Original'' to ``Original + Defect.'' This change appears as a slight change in the mode shown. We propose using small changes in observed modal motion to recover the locations and shapes of defects.}
    \label{fig:small_changes}
\end{figure}

As Fig.~\ref{fig:small_changes} illustrates, a small change in material properties (within a fixed geometry) results in a small change in modal motion. As most solid materials have Poisson's ratio $\approx0.3$~\cite{POPLAVKO201971}, the principal properties affecting motion are Young's modulus, which determines $K$, and density, which determines $M$. {\it Our  method is based on the insight that mode shapes on the surface of an object may reveal internal spatial inhomogeneities in these properties.}

\subsection{Challenge of Monocular Material Estimation}
\label{subsec:inferring}
We begin by setting up a simplified version of the inverse problem.
Assuming we perfectly measure all modes $u$ and frequencies $\omega$, then by Eq.~\ref{eq:geip}, we have the following minimization problem:
\setlength{\abovedisplayskip}{1.2pt}
\setlength{\belowdisplayskip}{1.2pt}
\begin{equation}
    K^*,M^* =\arg\min_{K,M} \left\lVert KU - M U\Lambda\right\rVert_2^2,
\label{eq:opt_K_M}
\end{equation}
where $U$ is the matrix whose columns are modes $u$, and $\Lambda$ is the diagonal matrix containing eigenvalues $\omega^2$. For a known geometry, this is a convex problem with respect to $K$ and $M$. However, Eq.~\ref{eq:opt_K_M} requires that we have access to all 3D modes and frequencies. In contrast, we will be working with experimentally-observed, image-space modes, incurring the following challenges:
\begin{enumerate}
    \itemsep-0.1em 
    \item \textit{Unseen degrees of freedom (DOFs).} We typically only observe a fraction of an object. For example, when observing a 3D cube with a monocular camera, one can see at most three of its sides, projected onto two directions of motion. Consider an 8x8x8 cubic mesh, which has $(8+1)^3=729$ vertices. With three directions of motion, it has $3\times 729=2187$ total DOFs. But a single monocular view of three sides of the cube can only observe $217$ vertices, moving in two directions of motion, amounting to $2\times 217=434$ image-space DOFs. This alone limits us to observing {\it fewer than 20\%} of the full-field DOFs for an 8x8x8 cube. 
    \item \textit{Unseen modes.} Theoretically, for discrete meshes, there are as many modes as there are DOFs. However, we can only capture modes at frequencies below the Nyquist sampling rate of the camera, which is $\text{FPS}/ 2$.
    \item \textit{Noise.} Aside from camera noise, there is noise from motion extraction, particularly in non-textured regions. 
\end{enumerate}

\begin{figure}[tb]
    \centering
    \includegraphics[width=0.45\textwidth]{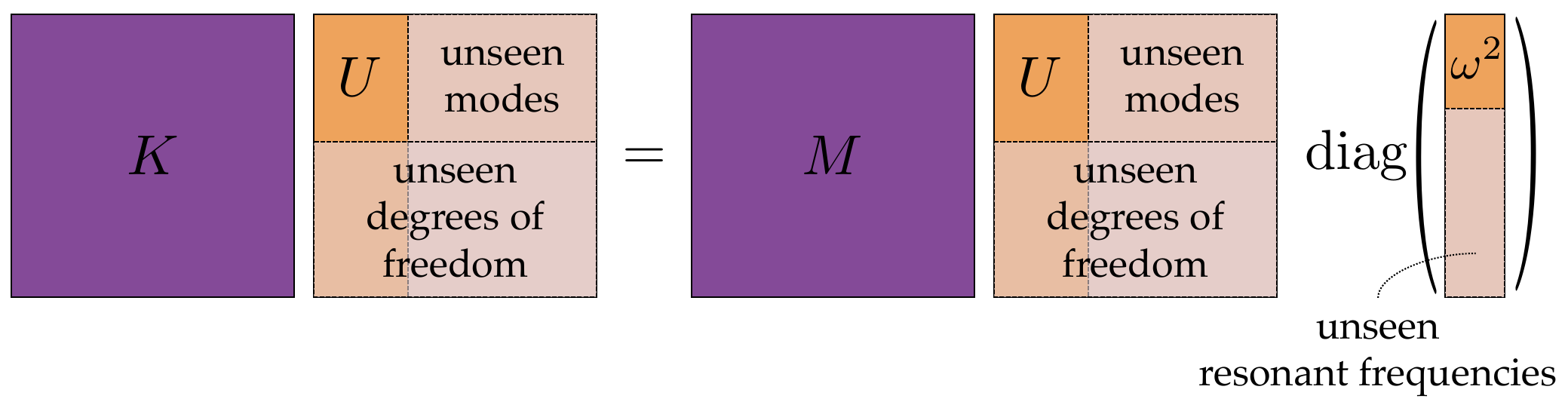}
    \caption{The generalized eigenvalue equation (Eq.~\ref{eq:geip}) defines the relationship between $K$, $M$ and $U$, $\mathbf{\omega^2}$. The matrix $U$ has columns corresponding to modes and rows corresponding to DOFs.
    The vector $\mathbf{\omega^2}$ contains associated eigenvalues. We would like to solve for $K$ and $M$ given partial information about $U$ and $\mathbf{\omega}^2$.}
    \label{fig:matrices}
    \vspace{-5mm}
\end{figure}

Due to limited data, the problem of solving for $K$ and $M$ (Eq.~\ref{eq:opt_K_M}) is ill-posed.\footnote{For a known geometry and complete mode and eigenvalue information, $K$ and $M$ are fully determined up to a scaling factor.}
As Fig.~\ref{fig:matrices} shows, observed data typically accounts for a tiny fraction of the matrices involved.

\section{Approach}
\label{sec:approach}
Our aim is to use motion features from a video to estimate material properties. This involves two stages: (1) motion extraction and image-space mode identification, and (2) solving for material properties that best match the observed image-space modes. The input is a video of an object, of which a mesh is known, and we assume that it is vibrating under linear elasticity (i.e., small motion). The output is two 3D images showing voxelized Young's modulus and density values throughout the object.

\subsection{Extracting Image-Space Modes from Video}
\label{subsec:motion_extraction}
\textbf{Motion extraction:} 
Since our approach relies on small, often imperceptible, motions, we need a way to extract sub-pixel motions from video.  
To quantify the displacements, we use the phase-based approach of Wadwha et al.~\cite{wadhwa2013phase}, which computes local phase shifts in a complex steerable pyramid~\cite{simoncelli1995steerable,simoncelli1992shiftable,portilla2000parametric}. This method has the advantage over other tracking methods (e.g., optical flow) of being robust to tiny motion, down to 0.001 pixel~\cite{davis2015visual}. The phase shifts are converted to pixel displacements using the approach proposed by Fleet and Jepson~\cite{fleet1990computation,wadhwa2017motion}.
To increase the signal-to-noise ratio, we filter out outlier pixels (i.e., top 1\% of displacement magnitudes) and then apply an amplitude-weighted Gaussian blur. The result of this step is a motion field for each frame, which quantifies the horizontal and vertical displacement at each pixel relative to the first frame.

\textbf{Identifying image-space modes:}
Modes are simply periodic motions occurring at particular frequencies, so we would expect them to appear as peaks in the power spectrum of motion amplitude. As is done in previous work that extracts image-space modes~\cite{davis2015image,davis2014visual, visvib2017pami}, we perform a discrete Fourier transform on the motion fields to analyze them in frequency space. To make this more concrete, let $\Delta x_t(x,y)$ and $\Delta y_t(x,y)$ be the horizontal and vertical displacements, respectively, at each pixel $(x,y)$ at frame $t$. The 1D FFT across time of these displacement fields results in complex-valued $\widehat{\Delta x}_\ell(x,y)$ and $\widehat{\Delta y}_\ell(x,y)$, corresponding to frequencies $f_\ell =(\text{FPS}\cdot\ell/T)$ Hz for $\ell\in[1,T]$. The motion power at frequency $f_\ell$ is then defined as $\left\lVert \begin{bmatrix} \widehat{\Delta x}_\ell, \widehat{\Delta y}_\ell \end{bmatrix} \right\rVert_2^2$ (dropping $(x,y)$ notation for clarity). A peak $\ell^*$ in the power spectrum ideally corresponds to a natural frequency $f_{\ell^*}$ and image-space mode given by $\begin{bmatrix}\text{Re}\left(\widehat{\Delta x}_{\ell^*}\right),\text{Re}\left(\widehat{\Delta y}_{\ell^*}\right)\end{bmatrix}$.

\textbf{Sampling image-space modes at mesh vertices:}
To approximate the 3D-to-2D projection matrix, a user manually identifies the pixel locations of several ``reference'' mesh vertices, and $P$ is the projection matrix that best maps the corresponding mesh coordinates to the image.
Using $P$, we map all of the mesh vertices from their 3D coordinates to 2D image coordinates. We then sample each image-space mode at the pixel locations of \textit{visible} mesh vertices. For mode $j$, we construct a vector $\gamma_j$ that contains the horizontal and vertical displacements of each mesh vertex at the corresponding natural frequency. Supposing we observe $q'$ out of $q$ mesh vertices, the vector $\gamma_j$ has the form
\begin{equation}
    \label{eq:gamma}
    \gamma_j=\left[\Delta x_1,\Delta y_1,\ldots, \Delta x_{q'},\Delta y_{q'}, 0, \ldots, 0\right]^\intercal\in\mathbb{R}^{2q},
\end{equation}
where $\Delta x_i$ is the horizontal (pixel) displacement and $\Delta y_i$ the vertical (pixel) displacement of vertex $i$. Unseen vertices are assigned displacements of 0, and for notational clarity, we position them at the end of the vector.

\subsection{Estimating Material Properties}
The matrices $K$ and $M$ are functions of Young's modulus and density. While typically expressed as \textit{global} matrices, they can be decomposed into \textit{local} matrices, which scale linearly with local material properties. As a result, $K$ and $M$ can each be written as a weighted sum of ``unit'' local matrices.
Specifically, we voxelize the volume containing the mesh so that each voxel contains a sub-collection of mesh elements. 
Given Young's modulus $w_e$ and density $v_e$ for each voxel, we express the global matrices as
\setlength{\abovedisplayskip}{0pt} \setlength{\abovedisplayshortskip}{0pt}
\setlength{\belowdisplayskip}{0pt} \setlength{\belowdisplayshortskip}{0pt}
\begin{equation}
\label{eq:K_M_constraints}
    K = \sum_{e=1}^m w_eK_e \text{\:\:\:and\:\:\:} M = \sum_{e=1}^m v_eM_e,
\end{equation}
where $K_e$ and $M_e$ are ``unit'' local stiffness and mass matrices, which we assemble using \texttt{FEniCS}~\cite{alnaes2015fenics}, and $m$ is the number of voxels.
This allows us to represent $K$ and $M$ as functions of
vectors $w,v\in\mathbb{R}^m$.
 
\subsubsection{Optimization Formulation}
\textbf{Data-matching objective:}
\label{sec:optimization}
Suppose we have $k$ modal observations, where $\widehat{\gamma}_i$ and $\widehat{\omega_i}$ are the $i$-th observed image-space mode and natural frequency, respectively. We would like to determine the voxel-wise Young's modulus values $w$ and density values $v$ that, when assembled into global stiffness and mass matrices, result in 3D modes $u_1,\ldots,u_k$ that agree with $\widehat{\gamma}_1, \ldots, \widehat{\gamma}_k$ when projected onto image-space. Since we do not know the full-field 3D modes, we need to include them as decision variables. Intuitively, the data-matching objective is to minimize $\left\lVert Pu_i - \widehat{\gamma}_i \right\rVert$ for each $i$.

\textbf{Regularization:} To make the solution well-defined, we choose to minimize total squared variation (TSV) of $w$ and $v$, which encourages spatial smoothness. Moreover, since we are estimating both stiffness and mass, the objective function can become arbitrarily low if we do not constrain the range of material-property values; this is because scaling $K$ and $M$ by a factor of $s$ still satisfies the generalized eigenvalue equation: $(sK) u = \omega^2 (sM) u$.
To resolve this ambiguity, we choose to minimize the deviation of $w$ from a mean value $\bar{w}$.
Regardless of $\bar{w}$, the relative differences in $w^*$, $v^*$ will not change, and for defect characterization, we generally only care about relative changes.

The resulting optimization problem is written as
\small
\begin{align}
\label{eq:opt}
    w^*, v^* &= \arg\hspace{-0.25in}\min_{\substack{ w,v\in\mathbb{R}^m \\ K,M\in\mathbb{R}^{n\times n}\\u_i\in\mathbb{R}^n,i=1,\ldots, k}} \hspace{-0.1in} \Bigg\{\frac{\alpha_u}{2k}\sum_{i=1}^k\left\lVert Pu_i-\widehat{\gamma}_i\right\rVert_2^2 \\ \notag
    &+ \frac{\alpha_w}{2m}\left\lVert \nabla^2 w \right\rVert_2^2 + \frac{\alpha_v}{2m}\left\lVert \nabla^2 v \right\rVert_2^2 + \left(\sum_{e=1}^mw_e/m -\bar{w}\right)^2 \Bigg\} \\ 
    \text{s.t. } & K = \sum_{e=1}^m w_e K_e, 
    \hspace{0.1in} M = \sum_{e=1}^m v_e M_e, \notag \\
    & K u_i = {\widehat{\omega}_i}^2 M u_i, \: i = 1,\ldots, k, \notag
\end{align}
\normalsize
where $\alpha_u$, $\alpha_w$, and $\alpha_v$ are hyperparameters that balance the objective terms. The effects of the regularization weights ($\alpha_w$ and $\alpha_v$) are discussed in the supplementary material. 

\subsubsection{Optimization Strategy}
As defined in Eq.~\ref{eq:dual}, we approximately solve Eq.~\ref{eq:opt} via a dual formulation of the problem.
The eigen-constraints in Eq.~\ref{eq:opt} are too strict to enforce directly, so we incorporate them as quadratic penalties in the dual problem. The weight of each penalty term is a dual variable, $y_i$, and we apply dual ascent to gradually increase these penalty weights.
\small
\begin{align}
\label{eq:dual}
    w^*, v^* &= \arg\hspace{-0.25in}\min_{\substack{ w,v\in\mathbb{R}^m \\ K,M\in\mathbb{R}^{n\times n} \\u_i\in\mathbb{R}^n,i=1,\ldots, k}} \hspace{-0.1in} \Bigg\{\frac{1}{2k}\sum_{i=1}^k y_i \left\lVert K u_i - {\widehat{\omega}_i}^2Mu_i \right\rVert_2^2 
    \\ \notag &+ \frac{\alpha_u}{2k}\sum_{i=1}^k\left\lVert Pu_i-\widehat{\gamma}_i\right\rVert_2^2 \\ \notag
    &+ \frac{\alpha_w}{2m}\left\lVert \nabla^2 w \right\rVert_2^2 + \frac{\alpha_v}{2m}\left\lVert \nabla^2 v \right\rVert_2^2 + \left(\sum_{e=1}^mw_e/m -\bar{w}\right)^2 \Bigg\} \\
    \text{s.t. } & K = \sum_{e=1}^m w_e K_e, 
    \hspace{0.1in} M = \sum_{e=1}^m v_e M_e. \notag
\end{align}
\normalsize

Eq.~\ref{eq:dual} is a non-convex problem, but it is quadratic with respect to $w,v$ for fixed $u_i$, and it is quadratic with respect to $u_i$ for fixed $w,v$. Our procedure is to iteratively compute the closed-form solution for $U=[u_1 \ldots u_k]$ and then $z=[w^\intercal, v^\intercal]^\intercal$, thereby minimizing the objective function at each step. We update the dual variables according to
\small
\begin{equation}
\label{eq:dual_update}
    y^{t+1}_i = y^t_i + \eta \left\lVert K^{t+1} u^{t+1}_i - {\widehat{\omega}}_i^2 M^{t+1} u^{t+1}_i \right\rVert_2,
\end{equation}
\normalsize
where $\eta>0$ is the dual-variable update rate. Once the decision variables have converged, we output the minimizing solution $z^*=[{w^*}^\intercal,{v^*}^\intercal]^\intercal$. $w^*$ and $v^*$ are the voxel-wise estimated Young's modulus and density values.
\section{Simulated Experiments}
\label{sec:simulated_experiments}
We test our approach on the simulated vibration of 3D cubes with ``defects," and we discuss the practical concerns of complex geometries, model mismatch, and damping.

\subsection{Creating Synthetic Data}

\textbf{Cube model:} We model a cube as a 10x10x10 hexahedral mesh, similar to a voxel grid. Each of the 1000 voxels is assigned a Young's modulus and density that correspond to either the primary material or a defect material. The material properties are chosen to resemble Jello and clay, respectively ($E_{\text{jello}} = 9000$ Pa, $\rho_{\text{jello}} = 1270$ kg/m3, $E_{\text{clay}} = 5\times 10^6$ Pa, $\rho_{\text{clay}} = 7620$ kg/m3). We set a homogeneous Poisson's ratio of $\nu=0.3$~\cite{POPLAVKO201971}.

\textbf{Vibration animation:} Once the cube's mesh and material properties have been defined, we run a transient analysis in COMSOL~\cite{comsol}, a commercial FEM software. The analysis calculates the cube's deformation over time given an initial condition. We choose an initial condition that mimics ``plucking'' a corner of the cube (e.g., an initial displacement vector of $(0.5,0.5,0.5)$ cm of the top-front corner) and keep the bottom surface fixed. The resulting simulation represents free vibration with a Dirichlet boundary condition. The simulation is 6 seconds long at 2000 FPS. From the calculated displacements, we create an animation of the cube deforming over time by plotting the motion of random points on the surface of the cube with \texttt{matplotlib}~\cite{matplotlib}.

\subsection{Implementation and Evaluation Details}
\textbf{Mode selection:} We use \texttt{scipy}'s~\cite{scipy} peak-finder to automatically identify peaks in the log-power spectrum of motion amplitude, as described in Sec.~\ref{subsec:motion_extraction}.
For a given simulation, this leads to around 20--30 selected peaks. Most peaks correspond to either a true mode or a linear combination of true modes whose frequencies fall in the same FFT frequency bin. However, a few peaks do not correspond to a true mode; we include these false modes in the synthetic results to best mimic analysis of real videos. 

\textbf{Inference cube mesh:} We infer on an 8x8x8 hexahedral mesh. Since the simulations are done with a 10x10x10 mesh, our results indicate robustness to a slight mesh mismatch. In the results presented, the simulation model and inference model use linear elements.

\textbf{Hyperparameters:} For every presented result from simulated data, $\alpha_w=10^{-10}$, $\alpha_v=10^{-7}$, and $\bar{w}=9000$. Keeping these hyperparameters fixed, we ran a hyperparameter search on a dataset of 12 cubes with various defects to identify good values for $\alpha_u$ (Eq.~\ref{eq:dual}) and $\eta$ (Eq.~\ref{eq:dual_update}). After testing all combinations of $\alpha_u\in\{1,10,100,1000\}$ and $\eta\in\{0.1,0.5,1,2,5,10\}$, we determined to set $\alpha_u=10$ if the number of input modes is $\geq10$; otherwise, $\alpha_u=1$.
The dual variables $y$ are always initialized to 1, with $\eta=1$. The decision variables $w$ and $v$ are initialized to homogeneous values of 9000 [Pa] and 1270 [kg/m3], respectively (the true values of the primary material). 

\textbf{Evaluation:}
Our method recovers \textit{relative changes} in material properties (see Sec.~\ref{sec:optimization}). As such, the normalized 3D images of estimated Young's modulus and density should match the true normalized properties. We use \textit{normalized correlation} between the estimated image and ground-truth image as the reconstruction score. Another way to assess estimated properties is to verify that they produce the same image-space modes and natural frequencies as the true properties (see supplementary material).

\begin{figure}[tb]
    \centering
    \includegraphics[width=0.47\textwidth]{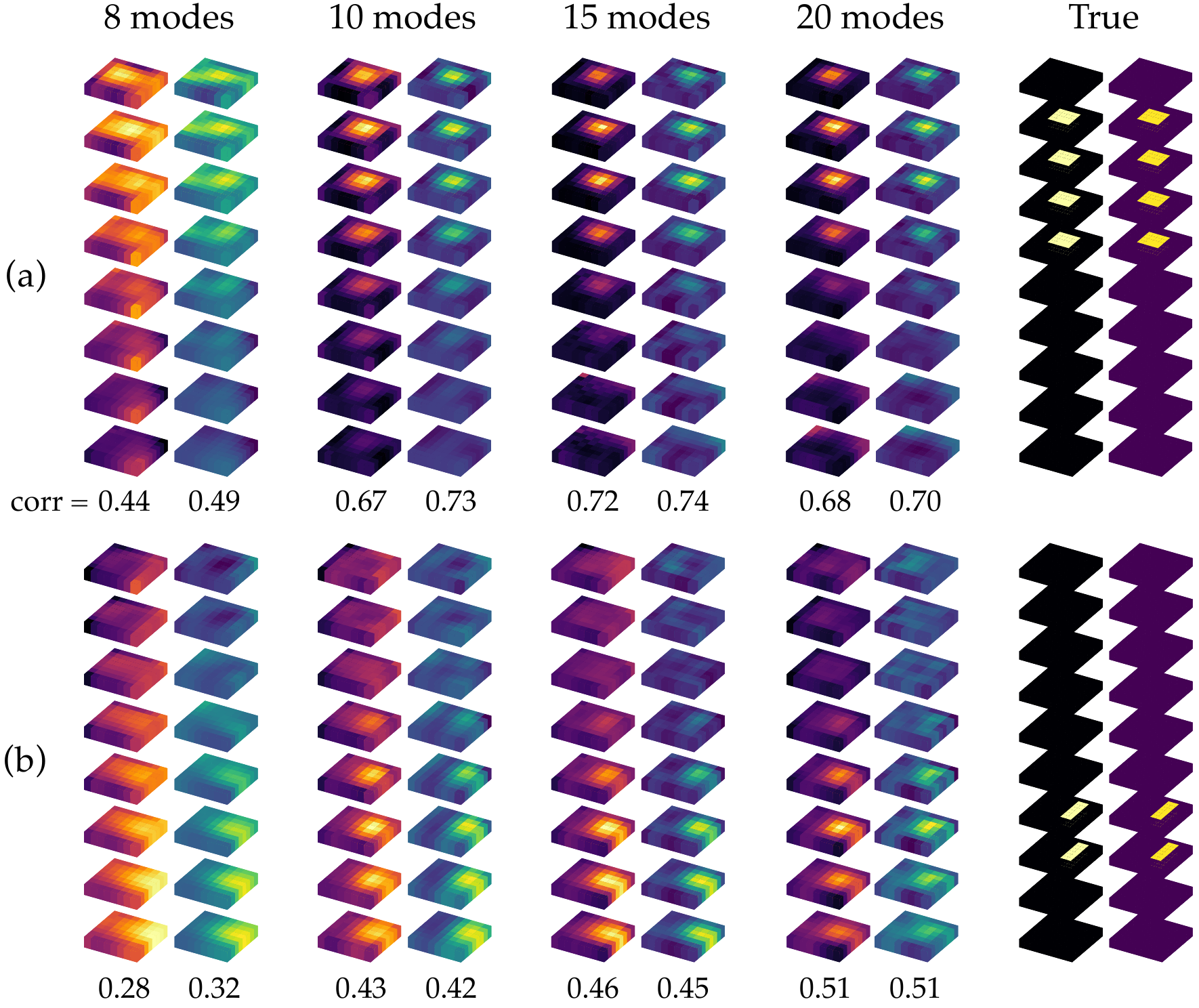}
    \caption{Reconstruction on two synthetic cubes with different defects. The given motion-extracted image-space modes range from the 8--20 lowest extracted modes. Normalized correlation generally increases as the number of modes increases. (b) is more challenging because the defect is smaller and closer to the bottom of the cube, where there is no motion.}
    \label{fig:nmodes}
    \vspace{-.2in}
\end{figure}

\subsection{Results}
Fig.~\ref{fig:nmodes} shows results for two different cubes with defects appearing at different locations. These results are obtained from noisy, motion-extracted image-space modes.
As more modes are observed, the inverse problem becomes better constrained, sharpening the image of the interior defect. Also note that a defect near the top of the cube is easier to identify than one near the bottom. This is because the base of the cube is fixed and thus provides less motion signal.

\begin{figure}[tb]
    \centering
    \includegraphics[width=0.475\textwidth]{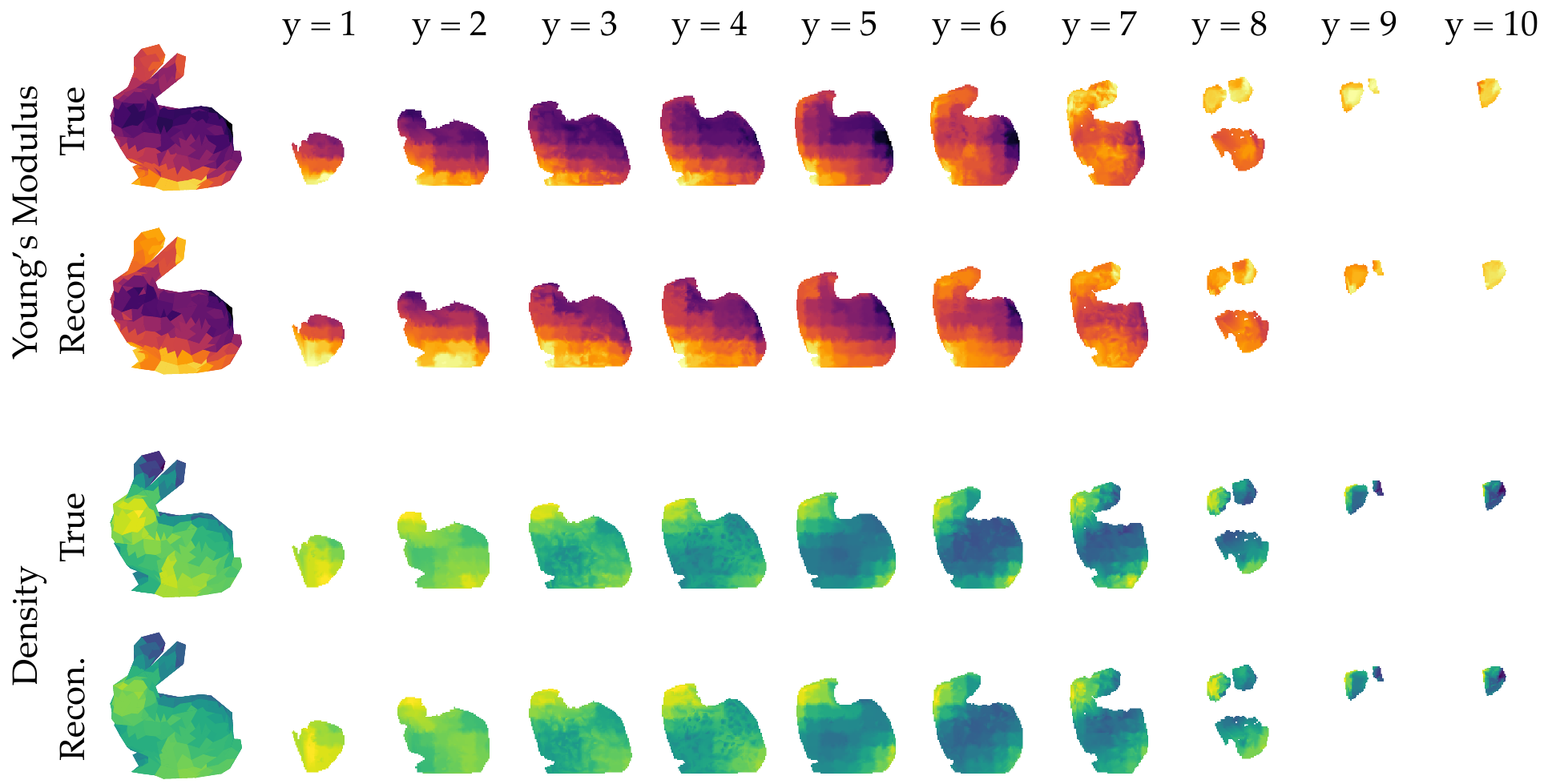}
    \caption{Reconstruction for the Stanford Bunny from (true) image-space modes. Slices along the $y$-axis are shown.}
    \label{fig:bunny}
    \vspace{-.2in}
\end{figure}

\textbf{Complex geometry:} To demonstrate the approach on a more complex geometry, Fig.~\ref{fig:bunny} shows a volumetric reconstruction of material properties for the Stanford Bunny~\cite{stanfordbunny}. We voxelize the volume containing the tetrahedral mesh\footnote{The bunny surface mesh is from {\scriptsize\url{https://www.thingiverse.com/thing:151081}}, and tetrahedralization is done with TetWild~\cite{tetwild}.} of the bunny into an 8x8x8 grid and match each mesh element to the nearest voxel, resulting in about 21 elements per voxel. 20 true image-space modes of the monocular view of the bunny shown are used for this reconstruction.

We next consider some challenges that may arise with real-world data: geometric mismatch and damping.
More investigations into model mismatch are provided in the supplementary material.

\vspace{-.1in}
\subsubsection{Geometric Mismatch}
\begin{figure}[ht]
    \centering
    \includegraphics[width=0.4\textwidth]{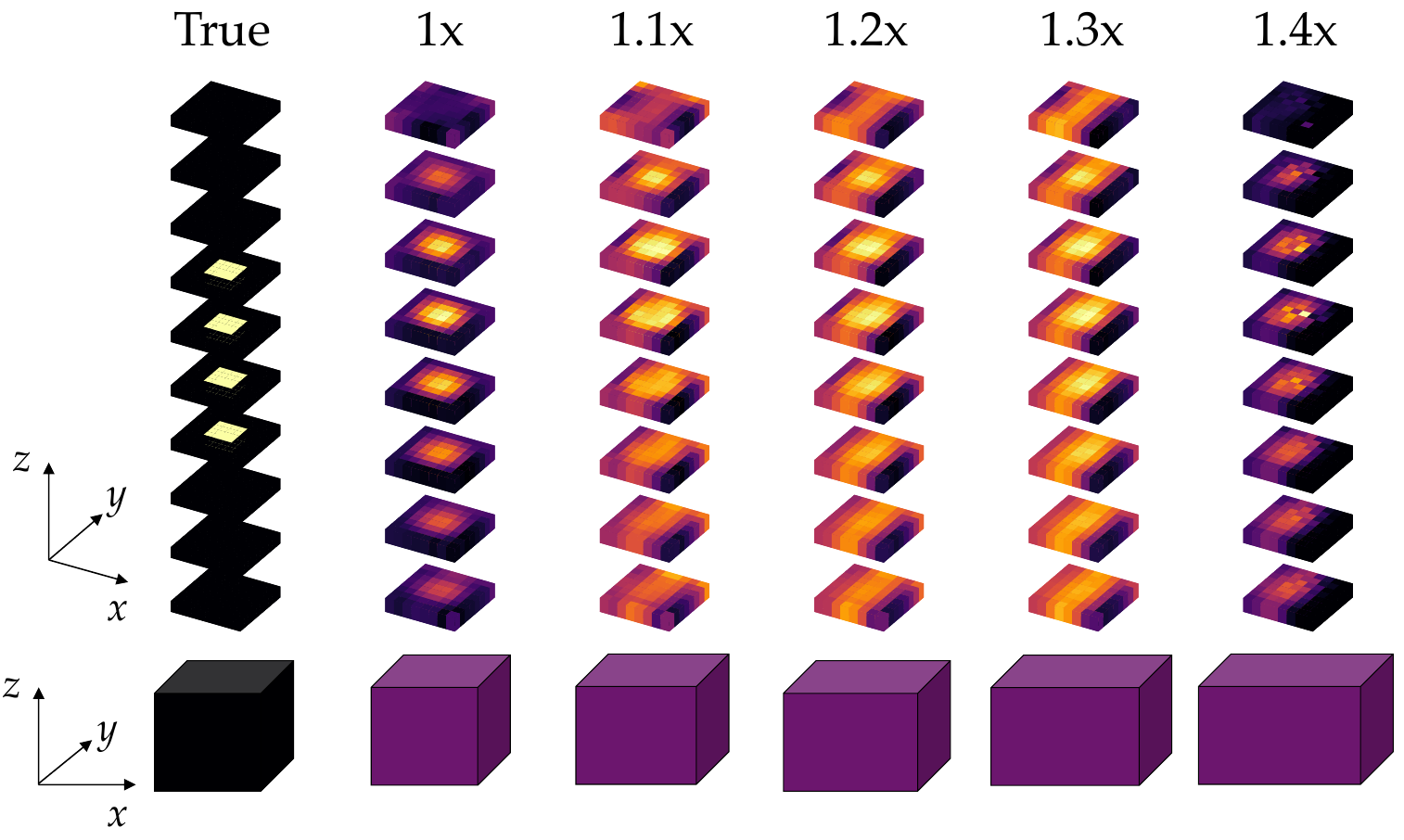}
    \captionsetup{type=figure}
    \caption{Geometric model mismatch. From 10 motion-extracted image-space modes, we infer on a mesh of various incorrect geometries, extending the inferred geometry width by a multiple of the true width. Results on Young’s modulus show that some geometric mismatch can be accommodated.}
    \label{fig:geo_mismatch}
    \vspace{-6mm}
\end{figure}

\vspace{-.1in}
Fig.~\ref{fig:geo_mismatch} shows what happens when the dimensions of the inference mesh do not match the cube's true dimensions. We gradually increase the length of the inferred cube geometry in the $x$-direction from 1 to 1.4 times the true length. Scaling the length in one direction results in a gradually degrading estimate of the defect size. However, even with 30\% geometric error, we are still able to distinguish that there is a defect located in the central region of the cube.

\subsubsection{Damping}
\label{subsubsec:damping}
\vspace{-.1in}
Real-world objects exhibit various types of damping, which can affect both the frequencies and relative phases of its modes. To simulate damping, we incorporate Rayleigh damping into our synthetic cubes.
Our damping parameters were estimated following the procedure outlined by Davis and Bouman et~al.~\cite{davis2015visual}, who fit a Lorentzian curve to a peak in the motion power spectrum to estimate the damping ratio. 
We find that Jello cubes exhibit significant damping: 
from a real video of one, we estimated critical damping ratios of 0.01749 at 12.5 Hz and 0.01999 at 15.5 Hz.

\begin{figure}[tb]
    \centering
    \includegraphics[width=0.47\textwidth]{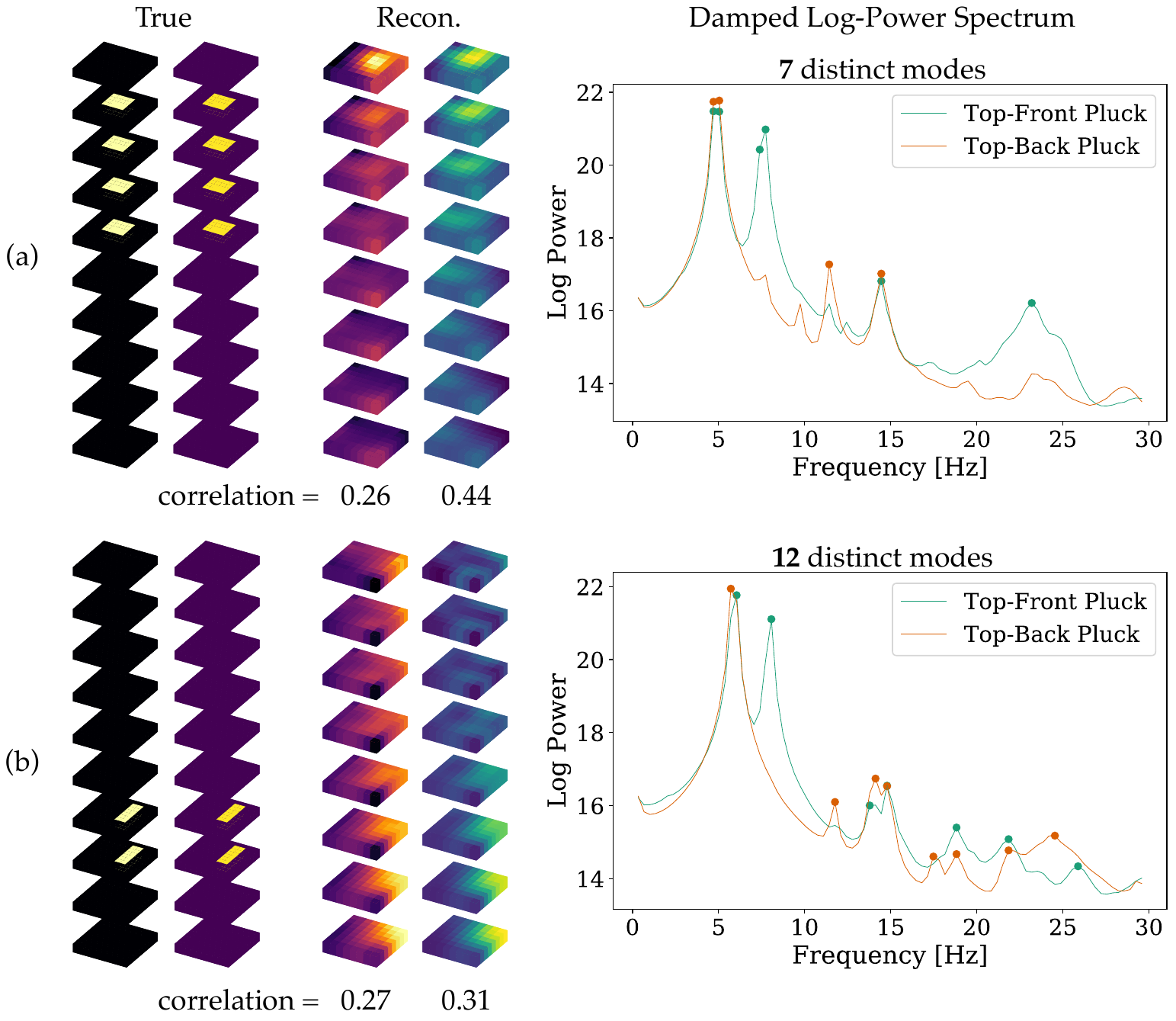}
    \caption{Reconstructions from two simulated damped cubes. Animations (3 seconds at 2000 FPS) of two different forcings are done: (1) a small initial displacement of the cube's top-front corner (``Top-Front Pluck") and (2) a small initial displacement of its top-back corner (``Top-Back Pluck''). Modes (marked as dots on the line plot) are selected based on the log-power spectrum of motion amplitude. Asymmetry plays a role in determining how many distinct modes are observable. As a cube becomes more asymmetric in its material-property distribution, its repeated eigenfrequencies become more separated. Since the defect in (b) is more off-center than the defect in (a), more distinct modes are identifiable in those simulations. In (a), with only 7 observed image-space modes, the reconstruction quality is consistent with Fig.~\ref{fig:nmodes}, which shows only a coarse defect reconstruction when given 8 modes. (Note: in (b), although the number of observed modes is $>10$, we show the reconstruction for $\alpha_u=1$ instead of 10.)}
    \label{fig:damping}
    \vspace{-5mm}
\end{figure}

\begin{figure*}[htb]
    \centering
    \includegraphics[width=0.76\textwidth]{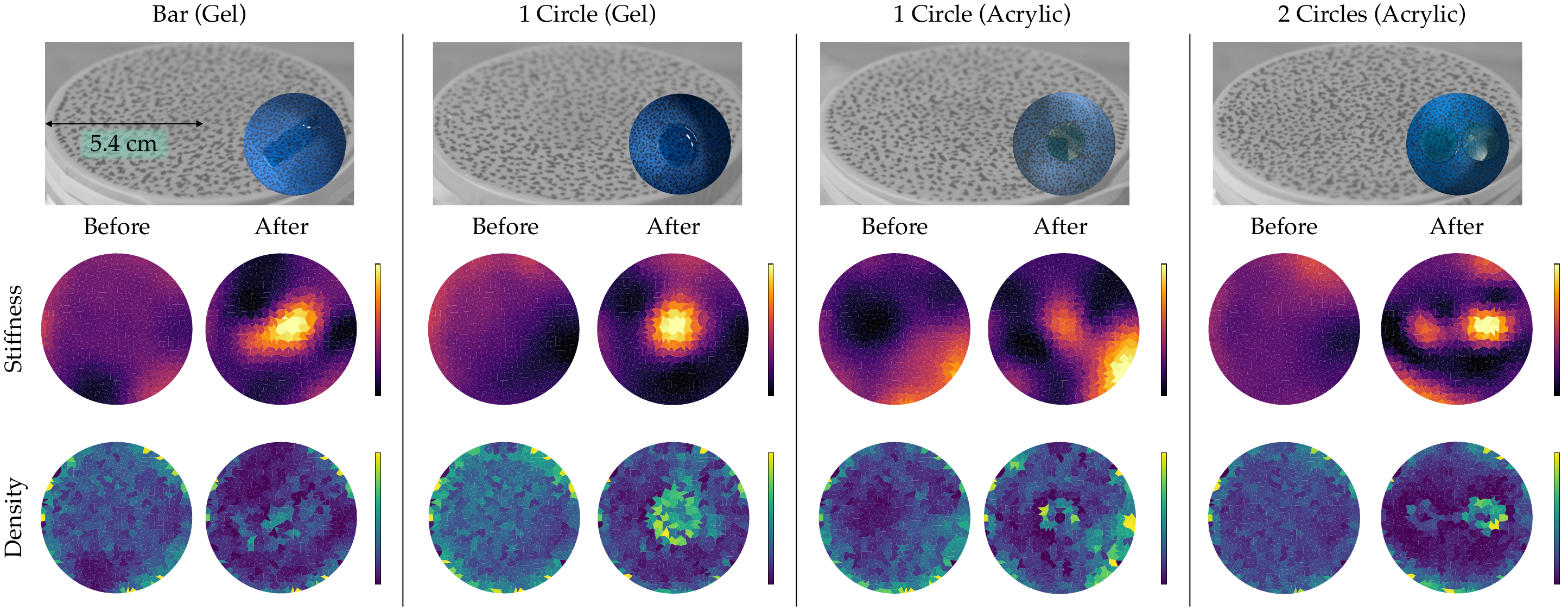}
    \caption{
        Reconstruction from real videos of drums. The defects shown are a gel bar, gel circle, acrylic circle, and two acrylic circles, applied to the underside of the drum head. For each defect, we recorded a video of the drum pre- and post-defect. One cannot see the defect in a video frame, but after applying our method, we were able to image the defects as changes in stiffness and density. For each type of defect, the ``Before'' and ``After'' material properties are plotted with the same normalized colormap.
        }
    \label{fig:drums}
    \vspace{-.25in}
\end{figure*}
\begin{figure}[ht]
    \centering
    \includegraphics[width=0.45\textwidth]{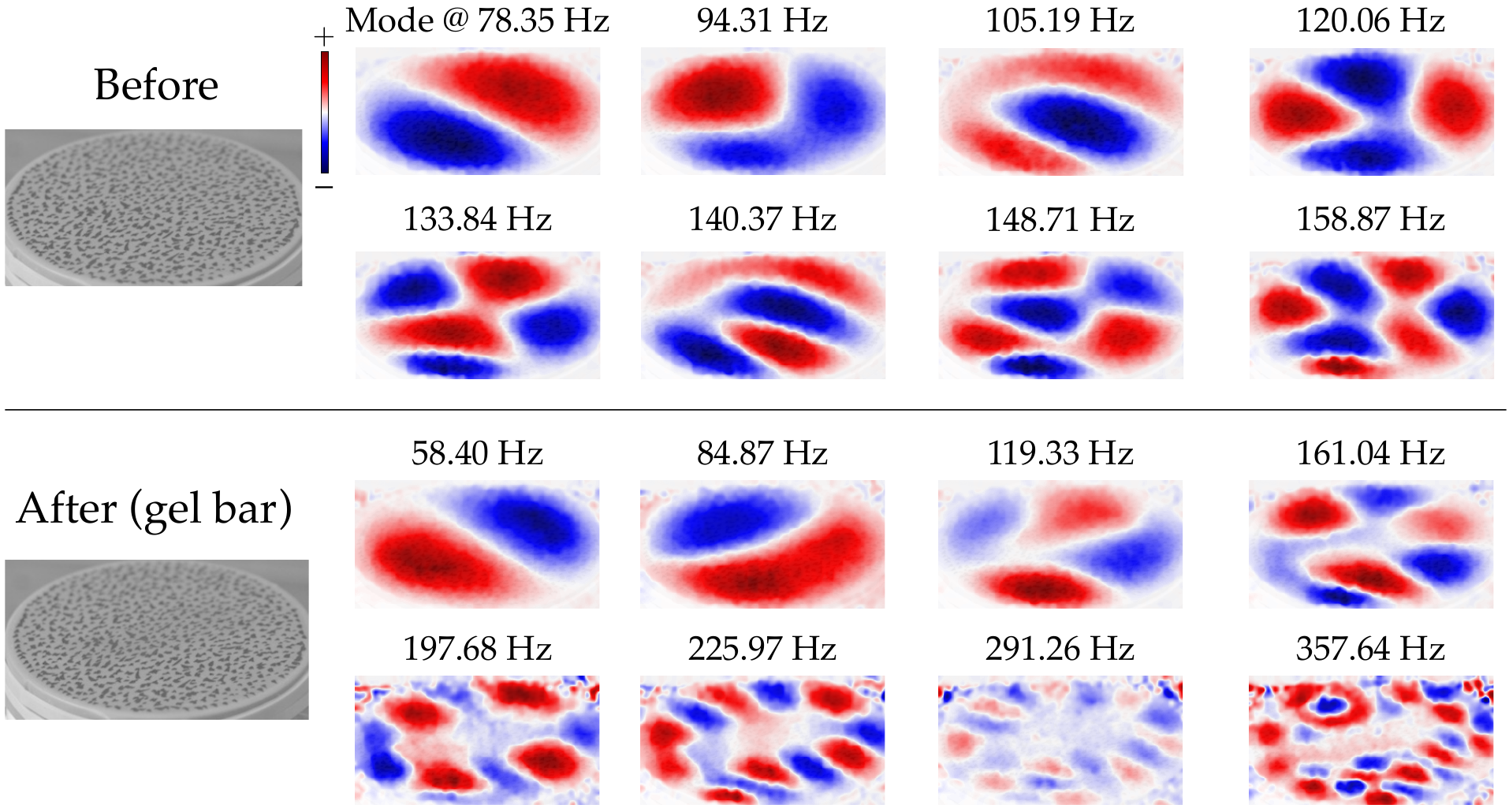}
    \caption{Extracted image-space modes from real videos of a drum, before and after a defect was introduced. The defect shown here is a gel rectangle, which was painted on the bottom of the drum head.
    Only vertical motion is shown.}
    \label{fig:drums_modes}
    \vspace{-.25in}
  \end{figure}

Through realistic simulations in COMSOL, we find that damping poses the additional challenge of fewer observable modes. We can increase the number of observed modes by extracting modes from multiple simulations with different initial conditions. For example, ``plucking" the top-back corner of a cube will cause slightly different modal expression than ``plucking" its top-front corner. Fig.~\ref{fig:damping} shows reconstruction for damped cubes. From two different plucking conditions, we are able to extract between 7--15 modes and use these modes to coarsely reconstruct the defect.

\section{Real-World Experiments}
\label{sec:real_experiments}
To demonstrate the potential of our approach in the real world, we applied it to real videos of drum heads and Jello cubes. With drum heads, we achieve reconstructions that allow one to discern distinct defect shapes, providing a proof-of-concept for defect discovery and characterization using our approach. 
The damping of 3D Jello cubes poses a challenge for extracting enough image-space modes for high-fidelity defect reconstruction; nonetheless, we are able to identify heterogeneity in the cube. 
Please refer to the supplementary material for details about the experiment setups, inference models, and hyperparameters.

\begin{figure*}[ht]
    \centering
    \includegraphics[width=0.99\textwidth]{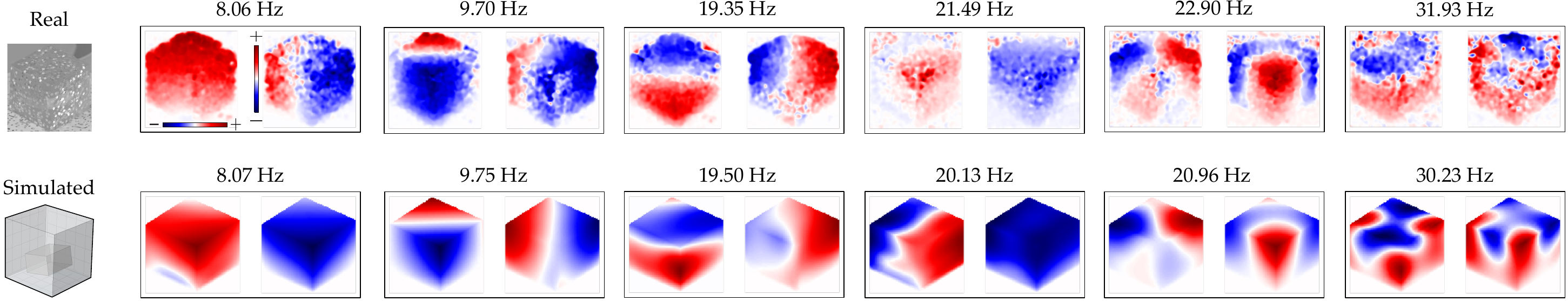}
    \captionsetup{type=figure}
    \caption{Extracted image-space modes from real videos of a Jello cube with an interior clay defect (``Real''). The true image-space modes identified from a COMSOL simulation of a cube with a defect are shown for comparison (``Simulated''). Each observed image-space mode has a corresponding simulated mode that appears similar in both image-space and eigenfrequency.
    }
    \label{fig:real_cubes_modes}
    \vspace{-.25in}
\end{figure*}

\begin{figure}[ht]
    \centering
    \includegraphics[width=0.475\textwidth]{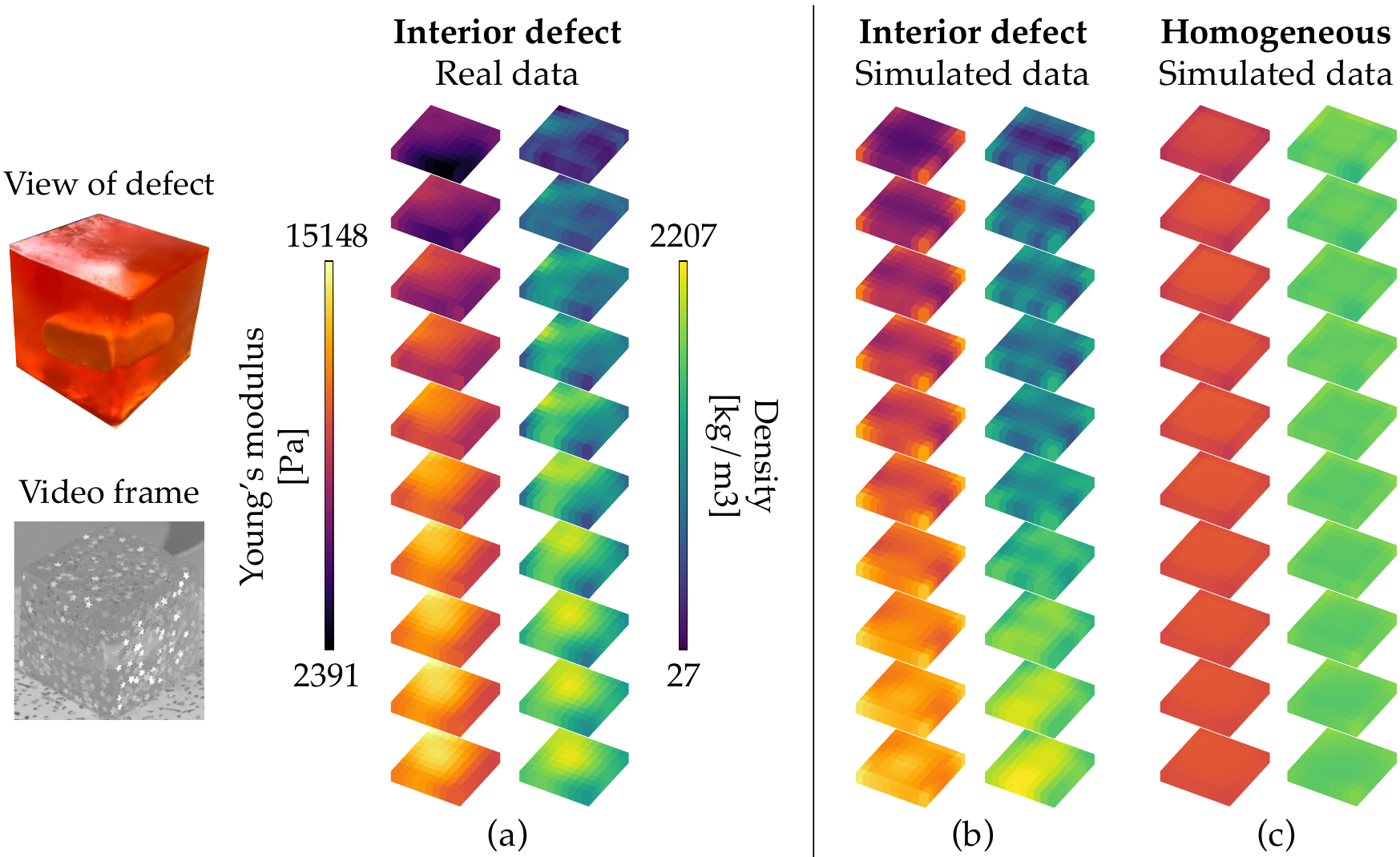}
    \captionsetup{type=figure}
    \vspace{-5mm}
    \caption{
        Reconstruction from real data vs. reconstructions from simulated data of the defect cube and a homogeneous cube. All reconstructions use 6 image-space modes and the same hyperparameters and are plotted with the same colormaps. As Fig.~\ref{fig:real_cubes_modes} shows, there is a one-to-one correspondence between the modes given for (a) and the modes given for (b).
        (a) is more similar to (b) than to (c), indicating that with 6 modes, we can differentiate between a cube with a defect and a homogeneous one.
        \vspace{-0.2in}
    }
    \label{fig:real_cubes_recon}
\end{figure}

\subsection{Real Drums}
We tested our method on a dataset of real drum heads, each altered with a defect beneath the surface. The defects were created from two materials: nail hardening gel (painted beneath the surface) or acrylic plastic circles (glued onto the bottom of the surface). Although all DOFs of the 2D membrane are visible in the video, solving for material properties is still ill-posed because we observe a limited number of projected modes (see Fig.~\ref{fig:drums_modes}).

\textbf{Results:} Fig.~\ref{fig:drums} shows estimated Young's moduli and densities for various drum heads, before and after defects were included. For both materials, the defect appears as a bright region in stiffness. Interestingly, gel and acrylic appear differently in their density estimations. For gel defects, there is a bright, filled region in the density map that corresponds to a higher mass from the defect. For acrylic defects, this change only appears on the edges of the defect. This is possibly because the acrylic circles are much stiffer than gel, which bends along with the rubber membrane.
These results indicate that our proposed approach could be used to identify not just the presence of a defect, but also its shape. 

\subsection{Real Cubes}
\label{sec:realcubes}
To gain further insight into practical challenges, we conducted an experiment on a real Jello cube with an interior clay defect. This object is more challenging than the drum membrane in two respects: (1) the high damping of Jello, perhaps due to its water content, and (2) the large proportion of unseen DOFs in the cube geometry. The cube had dimensions 4.9 x 4.7 x 4.5 cm, while the rectangular clay defect was of size 2.2 x 2.9 x 1.4 cm.
We recorded three videos of the cube under different initial deformation conditions (e.g., in one video, we lifted and then quickly released the top-front corner of the cube). Multiple videos allowed us to identify more unique modes and average duplicate ones. 

We created two COMSOL models that would be comparable to the real Jello cube: one simulated cube had a clay defect and the other did not. The Young's modulus values of the Jello and clay were set so that the natural frequencies would agree with those observed. The Rayleigh damping parameters were estimated following the method mentioned in Sec.~\ref{subsubsec:damping}. As illustrated in Fig.~\ref{fig:real_cubes_modes}, the COMSOL image-space modes of the simulated cube with a defect appear similar to those captured from the real Jello cube. 

\textbf{Results:} Fig.~\ref{fig:real_cubes_recon} shows the result of our approach applied to real video data of the Jello cube. 
The reconstruction is obtained using six unique, motion-extracted image-space modes.
As expected based on our findings in Figs.~\ref{fig:nmodes} and \ref{fig:damping}, we are able to recover only a large-scale estimation of material properties with six constraining modes. 
Still, it is very promising to have identified inhomogeneities in a real 3D object with our method. 
We further compare this real-data reconstruction to reconstructions obtained from simulated data of a homogeneous cube and one with a defect, showing that we achieve a solution that resembles the solution for the simulated defect cube more than it resembles the solution for the simulated homogeneous cube (Fig.~\ref{fig:real_cubes_recon}).

We have shown one example of attaining simulation quality on a real cube. Further work needs to be done to achieve consistent results across a variety of objects. Additional camera views are one plausible, simple solution.



\section{Limitations}

Our method assumes that materials are isotropic and linear elastic. Linear elasticity is only satisfied if the object's motion is small. Further, we assume that the geometry is, at least roughly, known ahead of time (see Fig.~\ref{fig:geo_mismatch}). 

For now, we have validated our method with a high-speed camera. We have not yet demonstrated the approach with consumer-grade cameras that bring additional challenges such as image compression and noise. Generally, the hardware required depends on the amplitude and frequencies of the modes. For large structures that vibrate below 100 Hz~\cite{MEMORY1995705,kolaini2018spacecraft,vibration2015sampaio}, a smartphone camera theoretically provides enough temporal frequency. Objects that vibrate more quickly require high-speed cameras. Tricks such as temporal aliasing via a strobe may expand the capabilities of a camera.


The primary challenge with applying this technique to real-world objects is capturing enough image-space modes to recover interior defects with high fidelity. 
Damping causes a reduction in the number of modes that can be extracted. 
We demonstrated that for damped Jello cubes, we could still recover some information from only six image-space modes (Sec~\ref{sec:realcubes}).
Even so, in many objects, damping will pose a more significant challenge.
In the future, acquiring more modal observations could be solved by exciting modes through mechanical vibration tables. 

\section{Conclusion}
\label{sec:discussion}

We have shown that it is possible to recover spatially-varying material properties of 3D objects from monocular video, even in regions unseen in the image. This can be done by decomposing 2D surface motion into image-space modes, and then solving for the Young's modulus and density values that agree with the observed modes. We demonstrated our method on synthetic and real-world data of objects ranging from 2D drum heads to a 3D bunny.  

Our results highlight that monocular videos are a simple, yet powerful, source of data for understanding the physical properties of objects around us. We believe that videos are a promising domain for further research into non-destructive testing, turning everyday visual sensors into tools for material characterization.
\section*{Acknowledgments}
The authors would like to thank Michael Rubinstein and Bill Freeman for their helpful discussions. This work is funded by BP. B.T.F. is supported by an NSF GRFP Fellowship and a Kortschak Scholarship. C.D. and A.C.O. acknowledge support from DOE award no. DE-SC0021358 and NSF award no. 1835735.

\clearpage
{\small
\bibliographystyle{ieee_fullname}
\bibliography{bibliography}
}

\end{document}


\title{Supplementary Material for Visual Vibration Tomography}
\maketitle
\pagestyle{plain}

\tableofcontents

\hypersetup{
    linkcolor=red
}
\appendix
\section{Optimization Runtime}
Recall that the optimal material properties and 3D modes are found by iteratively updating the solutions to the following optimization problem:
\small
\begin{align}
\label{eq:dual}
    w^*, v^* &= \arg\hspace{-0.25in}\min_{\substack{ w,v\in\mathbb{R}^m \\ K,M\in\mathbb{R}^{n\times n} \\u_i\in\mathbb{R}^n,i=1,\ldots, k}} \hspace{-0.1in} \Bigg\{\frac{1}{2k}\sum_{i=1}^k y_i \left\lVert K u_i - {\widehat{\omega}_i}^2Mu_i \right\rVert_2^2 
    \\ \notag &+ \frac{\alpha_u}{2k}\sum_{i=1}^k\left\lVert Pu_i-\widehat{\gamma}_i\right\rVert_2^2 \\ \notag
    &+ \frac{\alpha_w}{2m}\left\lVert \nabla^2 w \right\rVert_2^2 + \frac{\alpha_v}{2m}\left\lVert \nabla^2 v \right\rVert_2^2 + \left(\sum_{e=1}^mw_e/m -\bar{w}\right)^2 \Bigg\} \\
    \text{s.t. } & K = \sum_{e=1}^m w_e K_e, 
    \hspace{0.1in} M = \sum_{e=1}^m v_e M_e. \notag
\end{align}
\normalsize
For an 8x8x8 cube with linear hexahedral elements, one iteration takes 2--3 seconds (tested on an 8-core Intel Core i9, 32 GB RAM), and convergence usually happens within 100 iterations.


\section{Simulated Experiments}
\begin{figure}[ht]
    \centering
    \includegraphics[width=0.36\textwidth]{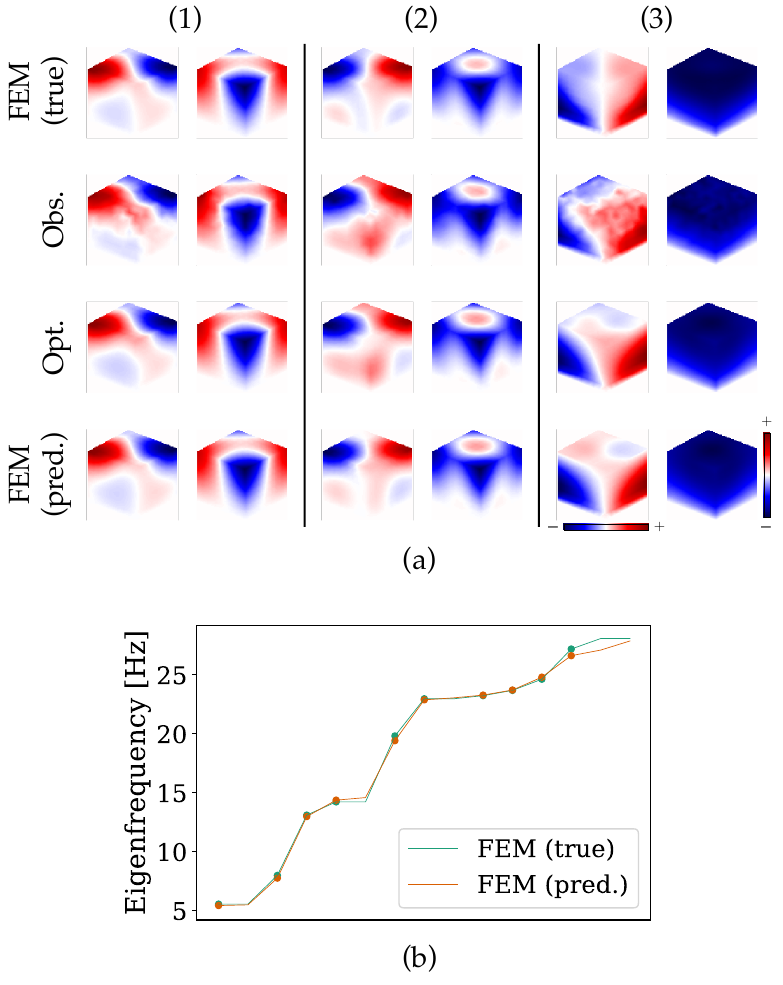}
    \caption{(a) Similarity of predicted image-space modes to true, observed, and optimized image-space modes for the cube sample shown in Fig.~\ref{fig:intrinsic_res}. ``Obs.'' mode is often a noisy version of ``FEM (true).'' ``Opt.'' refers to the optimized solution $U^*$ in Eq.~\ref{eq:dual}. ``FEM (pred.)'' is the image-space mode resulting from the estimated material properties. (b) Predicted eigenfrequencies vs. true eigenfrequencies. The frequencies of the 10 given motion-extracted image-space modes are marked by scatter dots.}
    \label{fig:forward_modes_full}
\end{figure}

\subsubsection{Predicted Image-Space Modes}
In addition to normalized correlation, a way to assess estimated material properties is to verify that they produce the same image-space modes and natural frequencies as the true properties.
Recall that 3D modes are a decision variable in our optimization scheme (Eq.~\ref{eq:dual}).
As Fig.~\ref{fig:forward_modes_full} shows, it is informative to compare the true FEM modes, observed modes, optimized modes, and predicted FEM modes, in image-space.
The optimization process usually de-noises observed modes. For some modes, spatially correlated noise may make it difficult to recover the true mode, but it is possible for the predicted FEM modes to still be similar to the truth (see example (2) in Fig.~\ref{fig:forward_modes_full}).

\subsubsection{Effect of Regularization} 
\begin{figure}[ht]
    \centering
    \includegraphics[height=1.5in]{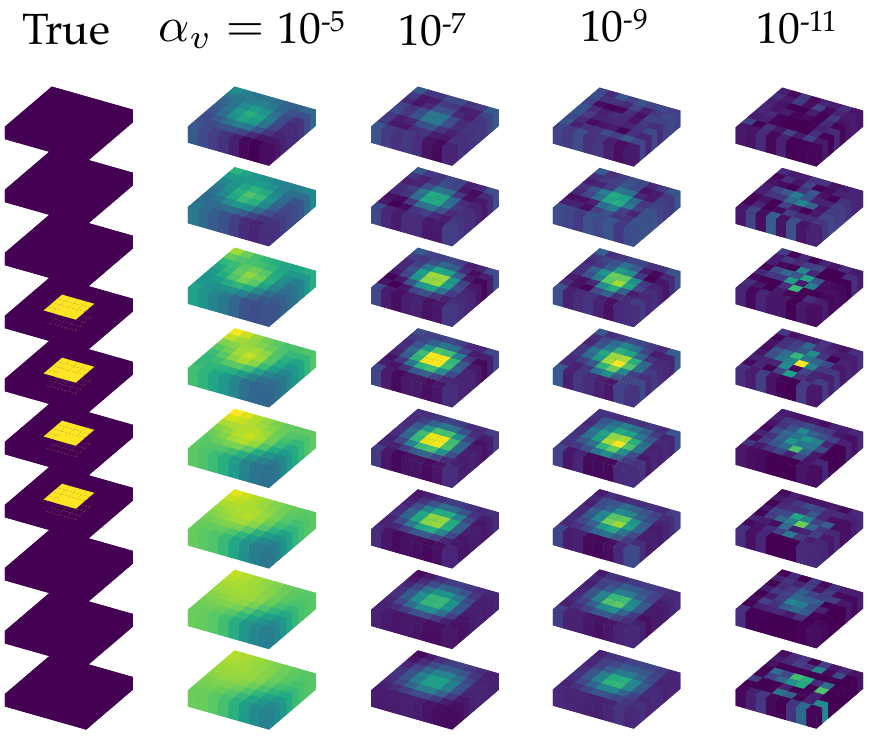}
    \caption{Effect of regularization. Here we show the estimated density for a cube sample (same as in Fig.~\ref{fig:intrinsic_res}). Keeping all else fixed, as $\alpha_v$ decreases, the image of the defect becomes crisper, but more sensitive to noise. Each estimation uses the same 10 motion-extracted image-space modes.}
    \label{fig:reg}
\end{figure}

The strength of spatial regularization on material properties affects the smoothness of the estimation. In Eq.~\ref{eq:dual}, the regularization weights for Young's modulus and density are $\alpha_w$ and $\alpha_v$, respectively. In Fig.~\ref{fig:reg}, we show how the estimated density image becomes sharper as $\alpha_v$ decreases (keeping all else fixed). This can result in a crisper picture of the defect, but can also make reconstruction more sensitive to noise.

\subsubsection{Intrinsic Resolution} 
\begin{figure}[ht]
    \centering
    \includegraphics[width=0.47\textwidth]{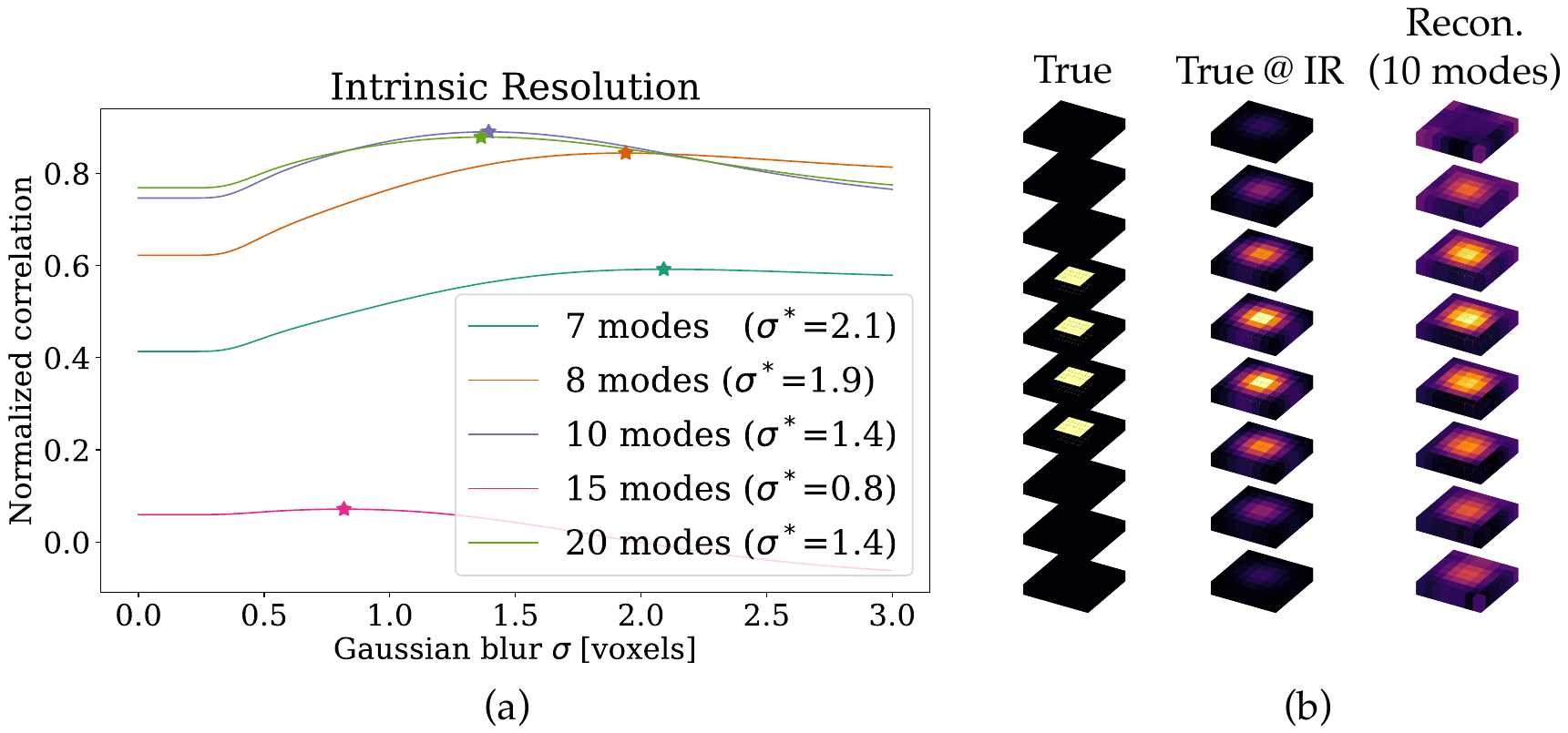}
    \caption{Intrinsic resolution of reconstructed volumes. Based on normalized correlation with the ground-truth material properties smoothed at different scales, one can approximate the intrinsic resolution of the reconstructed material properties. In (a), we plot normalized correlation versus Gaussian blur standard deviation $\sigma$, for the reconstruction of Young's modulus using different numbers of image-space modes (keeping all other hyperparameters fixed). As the number of observed modes increases, the reconstructed resolution also increases (i.e., smaller $\sigma$). (b) shows the true Young's modulus image blurred at the intrinsic resolution (IR) of the reconstruction given 10 image-space modes ($\sigma^*=1.4$ voxels).}
    \label{fig:intrinsic_res}
\end{figure}

Fig.~\ref{fig:intrinsic_res} demonstrates how, as the number of input image-space modes increases, the intrinsic resolution of the reconstruction improves.

\section{Simulated Experiments: Model Mismatch}
In the main material, we discussed geometry mismatch. Here, we investigate the effects of model mismatch in the Poisson's ratio and the mesh element order.

\subsubsection{Poisson's Ratio Mismatch}
\begin{figure}[ht]
    \centering
    \includegraphics[height=1.3in]{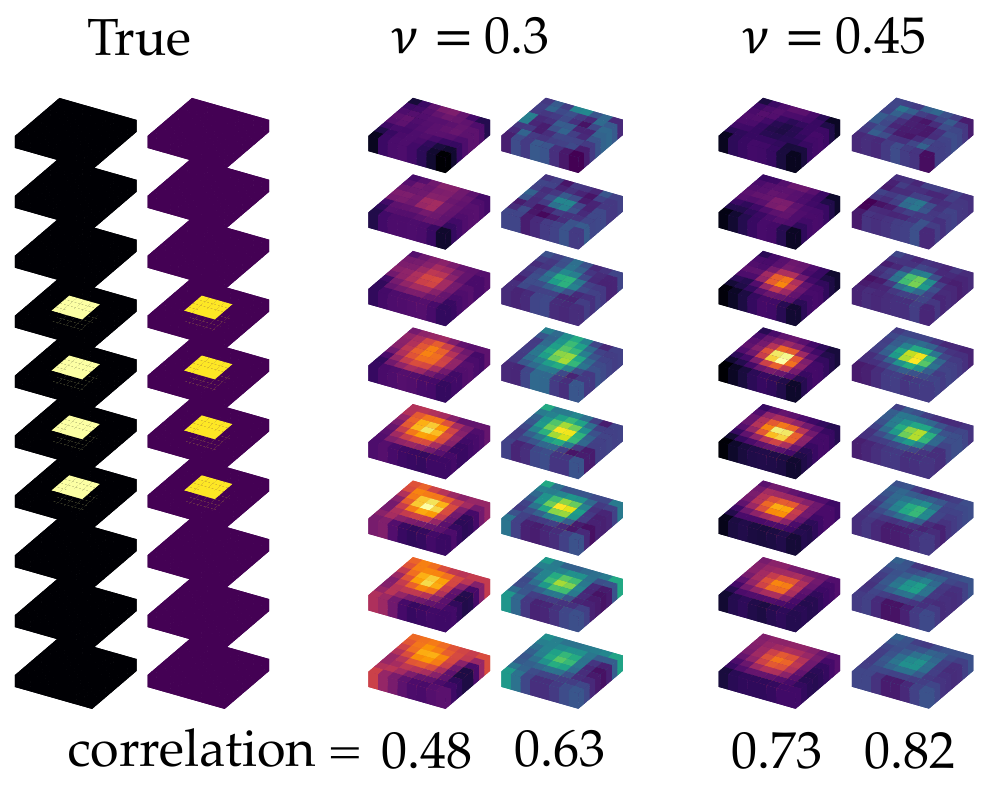}
    \vspace{-3mm}
    \caption{The effect of assuming a homogeneous Poisson's ratio. In the true cube, the main material has a Poisson's ratio of $\nu=0.45$, while the defect material has a Poisson's ratio of 0.3 (roughly corresponding to the values for Jello and clay, resp.). When inferring material properties, we assume a uniform Poisson's ratio across the entire cube. We find that this assumption does not hurt the reconstruction much, especially when $\nu$ is set to the Poisson's ratio of the main material. Both estimations use the same 20 motion-extracted image-space modes.}
    \label{fig:nu}
\end{figure}

The Poisson's ratio is a measure of the deformation of a material perpendicular to an applied force and ranges from 0.1 to 0.45~\cite{belyadi2019hydraulic}. Since our optimization formulation only estimates Young's modulus and density, we assume that every voxel has the same Poisson's ratio. Fig.~\ref{fig:nu} shows that this assumption does not significantly hurt the reconstruction of a defect, especially when assuming a Poisson's ratio that is closer to that of the main material. 

\subsubsection{Mesh Element Order Mismatch}
\begin{figure}[ht]
    \centering
    \includegraphics[width=0.45\textwidth]{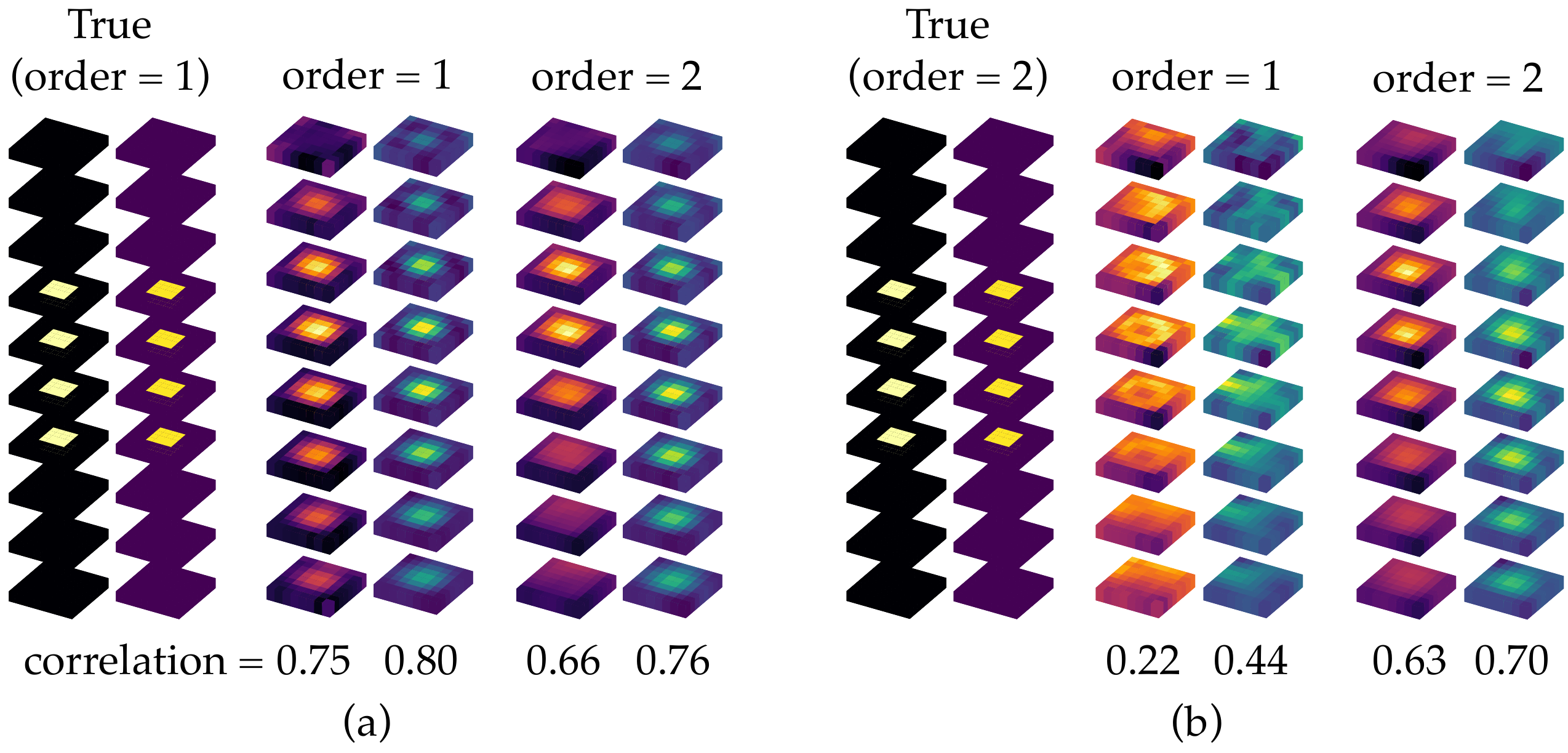}
    \vspace{-3mm}
    \caption{Element order mismatch. As (a) shows, when using linear elements in the forward model, any element order $\geq 1$ suffices in the inference model. However, in (b), reconstruction quality degrades when attempting to model a quadratic-element cube with linear elements.}
    \label{fig:element_order}
\end{figure}

In general, it is better to use higher-order elements to model a real-life object. However, there is a tradeoff in efficiency. A quadratic element approximates node displacements more accurately, but contains more DOFs than a linear element. Fig.~\ref{fig:element_order} shows what happens when the forward model uses quadratic elements, while the inference model uses linear elements. For real-world objects, one should choose the order that strikes the right balanace between approximation accuracy and computational cost.

\section{Drum Experiment}
\subsubsection{Drum Construction} The drums were constructed by fixing a thin rubber sheet over a 4"x4" PVC adaptor with a rubber band. We tested defects of two materials: nail hardening gel and acrylic plastic circles.
For each defect, we recorded a video of the homogeneous drum before the defect was applied for comparison. We drew a speckle pattern on the drum head for texture.

\subsubsection{Vibration-Capture Setup} 
\begin{figure}[h]
    \centering
    \includegraphics[height=1.5in]{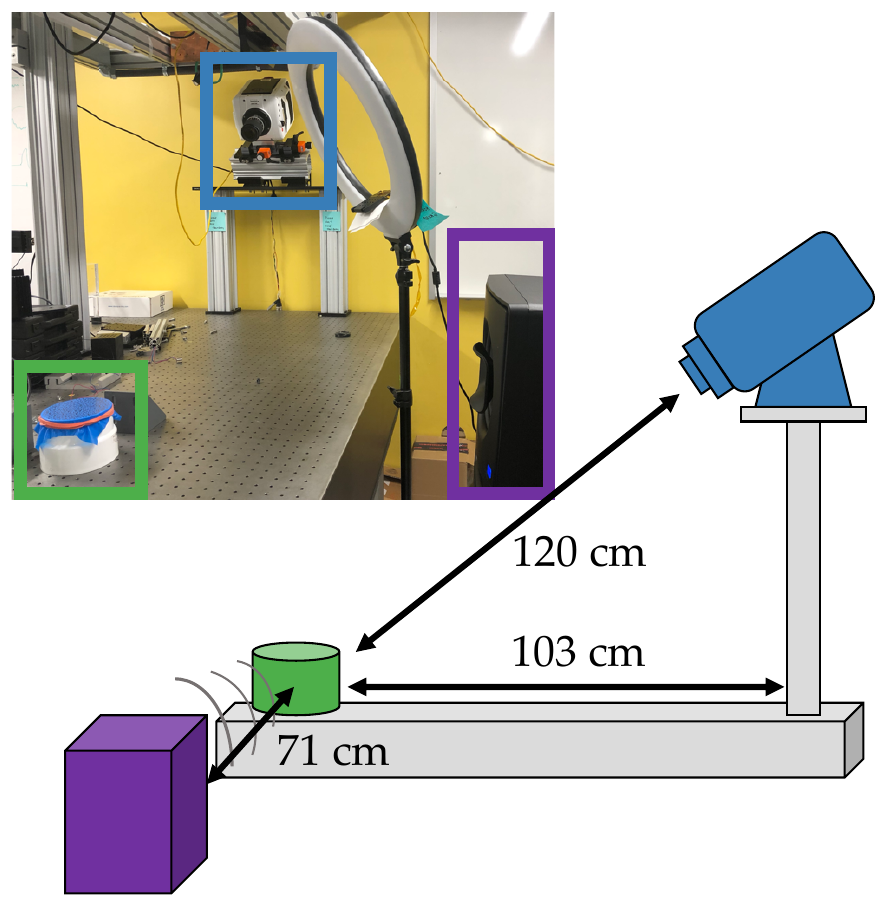} 
    \caption{Experimental setup for real drums. Vibrations were induced by a loudspeaker and recorded with a high-speed camera.}
    \label{fig:setup}
  \end{figure}

Fig.~\ref{fig:setup} shows a schematic of the setup. We taped the drum onto an optical table, with the high-speed camera standing on the same optical table. The excitation source was a PreSonus Sceptre S8 loudspeaker, which sat on a platform separate from the optical table and was pointed at the drum. For each video, we recorded the drum head's vibration in response to a 3.5-second linear frequency sweep (50--1000 Hz) played by the speaker.

\subsubsection{Video Capture} Our camera was a Phantom V1610 high-speed camera. Each video was captured at 6000 FPS at an image resolution of $288\times 384$. To reduce noise, we averaged every two frames for a resulting temporal frequency of 3000 FPS. Note that in Fig.~9 in the main material, the drums vibrate at frequencies below 120 Hz. While we chose to first demonstrate our approach using a high-speed camera, where compression and camera noise are less challenging, many modal frequencies can be captured on a consumer camera.

\subsubsection{Extracting Image-Space Modes} We found that in real videos, some level of manual selection was necessary to verify peaks in the motion amplitude spectrum as modal motion. For instance, spurious camera motion would often appear as spikes in the spectrum. Verification was done by visually inspecting the magnified motion in the video at the frequency in question (following the method proposed in \cite{wadhwa2013phase}). 
We believe that in the future this step could be automated.
The number of extracted modes ranged from 12 to 31, depending on the video. 

\subsubsection{Inference Details} We modeled each drum as a triangular membrane mesh with 1530 linear elements and inferred material properties on a 20x20 pixel grid. In the presented results, the hyperparameter values are $\alpha_u=10^{12}$, $\eta=1$, $\alpha_w=0.1$, $\alpha_v=0.1$, and $\bar{w}=10^6$. $w$ and $v$ are initialized to uniform values of $10^6$ [Pa] and $10^3$ [kg/m3], respectively, and reflect the estimated stiffness and density of latex.

\section{Jello Cube Experiment}
\subsubsection{Inference Details} Our inference model was a 10x10x10 hexadral mesh with linear elements. The optimization hyperparameters were $\alpha_u=0.1, \eta=1, \alpha_w=10^{-10}, \alpha_v=10^{-8},$ and $\bar{w}=10000$. $w$ and $v$ were initialized to $10000$ [Pa] and $1500$ [kg/m3], which are the estimated Young's modulus and measured density values of Jello.

{\small
\bibliographystyle{ieee_fullname}
\bibliography{bibliography}
}